\documentclass{article}

\usepackage{times}
\usepackage{graphicx} 
\usepackage{subfigure} 
\usepackage{dblfloatfix}

\usepackage{natbib}

\usepackage{algorithm}
\usepackage{algorithmic}

\usepackage{graphicx}
\usepackage{comment}
\usepackage{amsmath}
\usepackage{amssymb}
\usepackage{siunitx}
\usepackage{wrapfig}
\usepackage{todonotes}
\usepackage{enumitem}
\usepackage{siunitx}
\usepackage{dsfont}
\usepackage{xcolor}

\usepackage{hyperref}

\usepackage[accepted]{include/icml2017}

\newcommand{\myvec}[1]{\mathbf{#1}}
\newcommand{\myvecsym}[1]{\boldsymbol{#1}}

\newcommand{\vpi}{\myvecsym{\pi}}

\newcommand{\vtheta}{\myvecsym{\theta}}

\newcommand{\vtau}{\myvecsym{\tau}}

\newcommand{\vu}{\myvec{u}}

\newcommand{\E}{\mathbb{E}}

\newcommand{\be}{\begin{equation}}
\newcommand{\ee}{\end{equation}}
\newcommand{\bea}{\begin{eqnarray}}
\newcommand{\eea}{\end{eqnarray}}
\newcommand{\beaa}{\begin{eqnarray*}}
\newcommand{\eeaa}{\end{eqnarray*}}

\DeclareMathAlphabet{\mathpzc}{OT1}{pzc}{m}{n}

\newcommand\reffig[1]{Figure \ref{fig:#1}}

\DeclareMathOperator*{\argmax}{arg\,max}

\icmltitlerunning{Stabilising Experience Replay for Deep Multi-Agent Reinforcement Learning}
\begin{document}

\twocolumn[
\icmltitle{Stabilising Experience Replay for Deep Multi-Agent Reinforcement Learning}
\icmlsetsymbol{equal}{*}

\begin{icmlauthorlist}
  \icmlauthor{Jakob Foerster}{equal,oxf}
  \icmlauthor{Nantas Nardelli}{equal,oxf}
  \icmlauthor{Gregory Farquhar}{oxf} 
  \icmlauthor{Triantafyllos Afouras}{oxf} \\
  \icmlauthor{Philip. H. S. Torr}{oxf}
  \icmlauthor{Pushmeet Kohli}{msr}
  \icmlauthor{Shimon Whiteson}{oxf}
\end{icmlauthorlist}

\icmlaffiliation{oxf}{University of Oxford, Oxford, United Kingdom}
\icmlaffiliation{msr}{Microsoft Research, Redmond, USA}

\icmlcorrespondingauthor{Jakob Foerster}{jakob.foerster@cs.ox.ac.uk}
\icmlcorrespondingauthor{Nantas Nardelli}{nantas@robots.ox.ac.uk}

\icmlkeywords{Deep Learning, Reinforcement Learning, StarCraft, TorchCraft, Multi-Agent, Learning to Communicate}

\vskip 0.3in]
\printAffiliationsAndNotice{\icmlEqualContribution}

\begin{abstract}
\label{sec:abstract}

Many real-world problems, such as network packet routing and urban traffic
control, are naturally modeled as multi-agent \emph{reinforcement learning} (RL)
problems.  However, existing multi-agent RL methods typically scale poorly in
the problem size.  Therefore, a key challenge is to translate the success of
deep learning on single-agent RL to the multi-agent setting.  A major stumbling
block is that \emph{independent Q-learning}, the most popular multi-agent RL
method, introduces nonstationarity that makes it incompatible with the
\emph{experience replay memory} on which deep Q-learning relies.  This paper
proposes two methods that address this problem: 1) using a multi-agent variant
of importance sampling to naturally decay obsolete data and 2)  conditioning
each agent's value function on a \emph{fingerprint} that disambiguates the age
of the data sampled from the replay memory.  Results on a challenging
decentralised variant of \emph{StarCraft unit micromanagement}  confirm that
these methods enable the successful combination of experience replay with
multi-agent RL.

\end{abstract}
\section{Introduction}
\label{sec:intro}

\emph{Reinforcement learning} (RL), which enables an agent to learn control policies on-line given only sequences of observations and rewards, has emerged as a dominant paradigm for training autonomous systems. 
However,  many real-world problems, such as network packet delivery \citep{ye2015multi}, rubbish removal \citep{makar2001hierarchical}, and urban  traffic control \citep{kuyer:ecml08,VanDerPol16LICMAS}, are naturally modeled as cooperative multi-agent systems.  Unfortunately, tackling such problems with traditional RL is not straightforward.

If all agents observe the true state, then we can model a cooperative multi-agent system as a single meta-agent.  However, the size of this meta-agent's action space grows exponentially in the number of agents.  Furthermore, it is not applicable when each agent receives different observations that may not disambiguate the state, in which case decentralised policies must be learned.  

A popular alternative is \emph{independent Q-learning} (IQL) \citep{tan1993multi}, in which each agent independently learns its own policy, treating  other agents as part of the environment.   While IQL avoids the scalability problems of centralised learning, it introduces a new problem: the environment becomes nonstationary from the point of view of each agent, as it contains other agents who are themselves learning, ruling out any convergence guarantees.  Fortunately, substantial empirical evidence has shown that IQL often works well in practice \citep{matignon2012independent}.

Recently, the use of deep neural networks has dramatically improved the scalability of single-agent RL \citep{mnih2015human}.  However, one element key to the success of such approaches is the reliance on an \emph{experience replay memory}, which stores experience tuples that are sampled during training.  Experience replay not only helps to stabilise the training of a deep neural network, it also improves sample efficiency by repeatedly reusing experience tuples.
Unfortunately, the combination of experience replay with IQL appears to be  problematic: the nonstationarity introduced by IQL means that the dynamics that generated the data in the agent's replay memory no longer reflect the current dynamics in which it is learning.  While IQL without a replay memory can learn well despite nonstationarity so long as each agent is able to gradually track the other agents' policies, that seems hopeless with a replay memory constantly confusing the agent with obsolete experience.

To avoid this problem, previous work on deep multi-agent RL has limited the use of experience replay to short, recent buffers \citep{leibo2017multi} or simply disabled replay altogether \citep{foerster2016learning}.  However, these workarounds limit the sample efficiency and threaten the stability of multi-agent RL.  Consequently, the incompatibility of experience replay with IQL is emerging as a key stumbling block to scaling deep multi-agent RL to complex tasks.

In this paper, we propose two approaches for effectively incorporating experience replay into multi-agent RL. The first approach interprets the experience in the replay memory as  \emph{off-environment} data \citep{ciosek2017offer}.  By augmenting each tuple in the replay memory with the probability of the joint action in that tuple, according to the policies in use at that time, we can compute an importance sampling correction when the tuple is later sampled for training.  Since older data tends to generate lower importance weights, this approach naturally decays data as it becomes obsolete, preventing the confusion that a nonstationary replay memory would otherwise create.

The second approach is inspired by \emph{hyper Q-learning} \citep{tesauro2003extending}, which avoids the nonstationarity of IQL by having each agent learn a policy that conditions on an estimate of the other agents' policies inferred from observing their behaviour.  While it may seem hopeless to learn Q-functions in this much larger space, especially when each agent's policy is a deep neural network, we show that doing so is feasible as each agent need only condition on a low-dimensional \emph{fingerprint} that is sufficient to disambiguate where in the replay memory an experience tuple was sampled from.

We evaluate these methods on a decentralised variant of \emph{StarCraft unit micromanagement},\footnote{StarCraft and its expansion StarCraft: Brood War are trademarks of Blizzard Entertainment\texttrademark.}
 a challenging multi-agent benchmark  problem with a high dimensional, stochastic environment that exceeds the complexity of many commonly used multi-agent testbeds.  Our results confirm that, thanks to our proposed methods, experience replay can indeed be successfully combined with multi-agent $Q$-learning to allow for stable training of deep multi-agent value functions.
\section{Related Work}
\label{sec:related}

Multi-agent RL has a rich history \citep{busoniu2008comprehensive,yang2004multiagent} but has mostly focused on tabular settings and simple environments. The most commonly used method is independent Q-learning \citep{tan1993multi,MASfoundations09,Zawadzki:2014}, which we discuss further in Section \ref{sec:marl}.

Methods like hyper Q-learning \citep{tesauro2003extending}, also discussed in 
Section \ref{sec:marl}, and AWESOME \citep{conitzer2007awesome} try to tackle 
nonstationarity by tracking and conditioning each agent's learning process on 
their teammates' current policy, while \citet{da2006dealing} propose detecting 
and tracking different classes of traces on which to condition policy learning. 
\citet{kok2006collaborative} show that coordination can be learnt by estimating 
a global Q-function in the classical distributed setting supplemented with a 
coordination graph. In general, these techniques have so far not successfully 
been scaled to high-dimensional state spaces.

\citet{lauer2000algorithm} propose a variation of distributed Q-learning, a coordination-free method. However, they also argue that the simple estimation of the value function in the standard model-free fashion is not enough to solve multi-agent problems, and coordination through means such as communication \citep{mataric1998using} is required to ground separate observations to the full state function.

More recent work tries to leverage deep learning in multi-agent RL, mostly as a means to reason about the emergence of inter-agent communication.
\citet{tampuu2015multiagent} apply a framework that combines DQN with independent Q-learning to two-player pong. \citet{foerster2016learning} propose DIAL, an end-to-end differentiable architecture that allows agents to learn to communicate and has since been used by \citet{jorge2016learning} in a similar setting. \citet{sukhbaatar2016learning} also show that it is possible to learn to communicate by backpropagation. \citet{leibo2017multi} analyse the emergence of cooperation and defection when using multi-agent RL in mixed-cooperation environments such as the wolfpack problem. \citet{he2016opponent} address multi-agent learning by explicitly marginalising the opponents' strategy using a mixture of experts in the DQN.  Unlike our contributions, none of these papers directly aim to address the nonstationarity arising in multi-agent learning.

Our work is also broadly related to methods that attempt to allow for faster 
convergence of policy networks such as prioritized experience replay 
\citep{DBLP:journals/corr/SchaulQAS15}, a version of the standard replay memory 
that biases the sampling distribution based on the TD error. However, this 
method does not account for nonstationary environments and does not take 
into account the unique properties of the multi-agent setting.

\citet{wang2016sample} describe an importance sampling method for using off-policy experience in a single-agent actor-critic algorithm. However, to calculate policy-gradients, the importance ratios become products over potentially lengthy trajectories, introducing high variance that must be partially compensated for by truncation. By contrast, we address \emph{off-environment} learning and show that the multi-agent structure results in importance ratios that are simply products over the agents' policies.

Finally, in the context of StarCraft micromanagement,  \citet{usunier2016episodic} learn a centralised policy using standard single-agent RL. Their agent controls all the units owned by the player and observes the full state of the game. By contrast, we consider a decentralised task in which each unit has only partial observability.
\section{Background}
\label{sec:background}

We begin with background on single-agent and multi-agent reinforcement learning.

\subsection{Single-Agent Reinforcement Learning}

In a traditional RL problem, the agent aims to maximise its expected discounted return $R_t = \sum_{t=0}^{\infty}\gamma^t r_t$, where $r_t$ is the reward the agent receives at time $t$ and $\gamma \in [0,1)$ is the discount factor \citep{sutton1998reinforcement}. In a fully observable setting, the agent observes the true state of the environment $s_t \in S$, and chooses an action $u_t \in U$ according to a policy $\pi(u|s)$.

The action-value function $Q$ of a policy $\pi$ is $Q^{\pi}(s,u) = \E \left[R_t | s_t = s,u_t = u \right]$. The Bellman optimality equation,
\begin{equation}
\begin{split}
Q^*(s,u) &= \mathcal{T}Q^*(s,u) \\
&= r(s,u) +
\gamma \sum_{s'} P(s'|s,u) \max_{u'} Q^*(s', u'),
\end{split}
\end{equation}
recursively represents the  optimal $Q$-function $Q^{*}(s,u) = \max_{\pi} Q^{\pi}(s,u)$ as a function of the expected immediate reward $r(s,u)$ and the transition function $P(s'|s,u)$, which in turn yields an optimal greedy policy $\pi^*(u|s) = \delta(\argmax_{u'} Q(s,u')- u)$. $Q$-learning \citep{watkins1989learning} uses a sample-based approximation of $\mathcal{T}$ to iteratively improve the $Q$-function.
In deep $Q$-learning \citep{mnih2015human}, the $Q$-function is represented by a
neural network parameterised by $\theta$. During training, actions are chosen at each
timestep according to an exploration policy, such as an $\epsilon$-greedy policy that
selects the currently estimated best action $\argmax_uQ(s,u)$ with probability
$1-\epsilon$, and takes a random exploratory action with probability $\epsilon$. The reward
and next state are observed, and the tuple $\langle s,u,r,s'\rangle$ is stored in a \emph{replay memory}. The parameters $\theta$ are learned by sampling batches of $b$ transitions from the replay memory, and minimising the squared TD-error:
\begin{equation}
\mathcal{L}(\theta) = \sum_{i=1}^b [( y_i^{DQN} - Q(s,u; \theta))^{2}],
\label{eq:loss}
\end{equation}
with a target $y_i^{DQN} =  r_i + \gamma \max_{u_i'} Q(s_i',u_i';\theta^{-})$, where  $\theta^{-}$ are the parameters of a target network periodically copied from $\theta$ and frozen for a number of iterations. The replay memory stabilises learning,  prevents the network from overfitting to recent experiences, and improves sample efficiency.
In partially observable settings, agents must in general condition on their entire action-observation history, or a sufficient stastistic thereof.  In  deep RL, this is accomplished by modelling the $Q$-function with a recurrent neural network \citep{hausknecht2015deep}, utilising a gated architecture such as LSTM \citep{hochreiter1997long} or GRU \citep{chung2014empirical}.


\subsection{Multi-Agent Reinforcement Learning}
\label{sec:marl}

We consider a fully cooperative multi-agent setting in which $n$ agents
identified by $a \in A \equiv \{1,...,n\}$ participate in a stochastic game, $G$,
described by a tuple $ G = \langle S, U, P, r, Z, O, n, \gamma\rangle$.
The environment occupies states $s \in S$, in which, at every time step, each agent takes an action $u_a \in U$, forming a joint action $\mathbf{u} \in \mathbf{U} \equiv U^{n}$. 
State transition probabilities are defined by $P(s'|s,\mathbf{u}): S \times \mathbf{U} \times S \rightarrow [0,1]$. As the agents are fully cooperative, they share the same reward function $r(s,\mathbf{u}): S \times \mathbf{U} \rightarrow \mathbb{R}$. 

Each agent's observations  $z \in Z$ are governed by an observation function $O(s, a): S \times A \rightarrow Z $. For notational simplicity, this observation function is deterministic, i.e., we model only perceptual aliasing and not noise.  However, extending our methods to noisy observation functions is straightforward.
Each agent $a$ conditions its behaviour on its own action-observation history $\tau_a \in T \equiv (Z \times U)^{*}$, according to its policy $\pi_a(u_a|\tau_a): T \times U \rightarrow [0,1]$. After each transition, the action $u_a$ and new observation $O(s,a)$ are added to $\tau_a$, forming $\tau_a'$. We  denote joint quantities over agents in bold, and joint quantities over agents other than $a$ with the subscript ${-a}$, so that, e.g., $\mathbf{u} = [u_a, {\vu_{-a}}]$.

In \emph{independent Q-learning} (IQL) \citep{tan1993multi}, the simplest and most popular approach to multi-agent RL, each agent learns its own Q-function that conditions only on the state and its own action.  Since our setting is partially observable, IQL can be implemented by having each agent condition on its action-observation history, i.e., $Q_a(\tau_a,u_a)$.  In deep RL, this can be achieved by having each agent perform DQN using a recurrent neural network trained on its own observations and actions.

IQL is appealing because it avoids the scalability problems of trying to learn a joint Q-function that conditions on $\mathbf{u}$, since $|\mathbf{U}|$ grows exponentially in the number of agents.  It is also naturally suited to partially observable settings, since, by construction, it learns decentralised policies in which each agent's action conditions only on its own observations.  

However, IQL introduces a key problem: the environment becomes nonstationary from the point of view each agent, as it contains other agents who are themselves learning, ruling out any convergence guarantees.  On the one hand, the conventional wisdom is that this problem is not severe in practice, and 
substantial empirical results have demonstrated success with IQL \citep{matignon2012independent}.  On the other hand, such results do not involve deep learning.  

As discussed earlier, deep RL relies heavily on experience replay and the combination of experience replay with IQL appears to be  problematic: the nonstationarity introduced by IQL means that the dynamics that generated the data in the agent's replay memory no longer reflect the current dynamics in which it is learning.  While IQL without a replay memory can learn well despite nonstationarity so long as each agent is able to gradually track the other agents' policies, that seems hopeless with a replay memory constantly confusing the agent with obsolete experience.  In the next section, we propose methods to address this problem.
\section{Methods}
\label{sec:methods}

To avoid the difficulty of combining IQL with experience replay, previous work
on deep multi-agent RL has limited the use of experience replay to short, recent
buffers \citep{leibo2017multi} or simply disabled replay altogether
\citep{foerster2016learning}.
However, these workarounds limit the sample efficiency and threaten the
stability of multi-agent RL. In this section, we propose two approaches for
effectively incorporating experience replay into multi-agent RL.

\subsection{Multi-Agent Importance Sampling}

\begin{figure*}[t]
	\includegraphics[width=\textwidth]{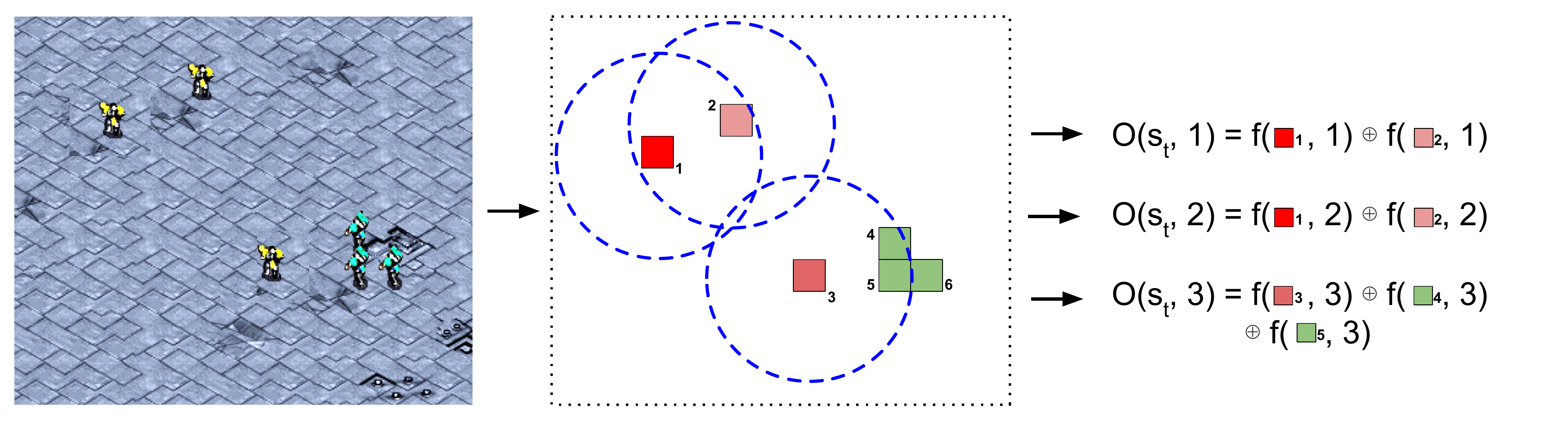}
	\caption{An example of the observations obtained by all agents at each time 
    step $t$. The function f provides a set of features for each unit in the
    agent's field of view, which are concatenated. The feature set is
    \{\texttt{distance}, \texttt{relative x}, \texttt{relative y},
    \texttt{health points}, \texttt{weapon cooldown}\}. Each quantity is
    normalised by its maximum possible value.}
	\label{fig:problem-setting}
\end{figure*}

We can address the non-stationarity present in IQL by developing an importance
sampling scheme for the multi-agent setting.  Just as an RL agent can use
importance sampling to learn \emph{off-policy} from data gathered when its own
policy was different, so too can it learn \emph{off-environment}
\citep{ciosek2017offer} from data gathered in a different environment. Since IQL
treats other agents' policies as part of the environment, off-environment
importance sampling can be used to stabilise experience replay. In particular,
since we know the policies of the agents at each stage of training, we know
exactly the manner in which the environment is changing, and can thereby correct
for it with importance weighting, as follows. We consider first a
fully-observable multi-agent setting. If the $Q$-functions can condition
directly on the true state $s$, we can write the Bellman optimality equation for
a single agent given the policies of all other agents:
\begin{multline}
Q^*_a(s,u_a|\vpi_{-a}) = \sum_{\vu_{-a}} \vpi_{-a}(\vu_{-a}|s) \bigg[ r(s, u_a,
\vu_{-a}) + \\
\gamma \sum_{s'} P(s'|s,u_a,\vu_{-a}) \max_{u_a'} Q^*_a(s', u_a') \bigg].
\label{eq:fully_obs_bellman}
\end{multline}
The nonstationary component of this equation is $\vpi_{-a}(\vu_{-a}|s) = \Pi_{i
\in
    -a}\pi_i(u_i|s)$, which changes as the other agents' policies change over
    time. Therefore, to enable importance sampling, at the time of collection
    $t_c$, we record $\vpi_{-a}^{t_c}(\vu_{-a}|s) $ in the replay memory,
    forming an augmented transition tuple ${\langle
    s,u_a,r,\pi(\vu_{-a}|s),s'\rangle}^{(t_c)}$.

At the time of replay $t_r$, we train off-environment by minimising an
importance weighted loss function:
\begin{equation}
\mathcal{L}(\theta) = \sum_{i=1}^b 
\frac{\vpi_{-a}^{t_r}(\vu_{-a}|s)}{\vpi_{-a}^{t_i}(\vu_{-a}|s)}[( y_i^{DQN} - 
Q(s,u; \theta))^{2}],
\label{eq:is-loss}
\end{equation}
where $t_i$ is the time of collection of the $i$-th sample.

The derivation of the non-stationary parts of the Bellman equation in the
partially observable multi-agent setting is considerably more complex as the
agents' action-observation histories are correlated in a complex fashion that
depends on the agents' policies as well as the transition and observation
functions.

To make progress, we can define an augmented state space $\hat{s} = \{s,
\vtau_{-a}\} \in \hat{S} = S \times  T^{n-1}$. This state space includes both 
the original state $s$ and the action-observation history of the other agents
$\vtau_{-a}$. We also define a corresponding observation function $\hat{O}$
such that $\hat{O}(\hat{s},a) = O(s,a)$. With these definitions in place, we define a
new reward function $\hat{r}(\hat{s},u) = \sum_{\vu_{-a}}
\vpi_{-a}(\vu_{-a}|\vtau_{-a}) r(s,\vu)$ and a new transition function,
\begin{multline}
\hat{P}(\hat{s}'|\hat{s},u) = P(s', \tau'|s,\tau,u) = \\ 
\sum_{\vu_{-a}} \vpi_{-a}(\vu_{-a}|\vtau_{-a}) P(s'|s,\vu) 
p(\vtau_{-a}'|\vtau_{-a},\vu_{-a},s').
\end{multline}
All other elements of the augmented game $\hat{G}$ are adopted from the
original game $G$. This also includes $T$, the space of action-observation
histories. The augmented game is then specified by $\hat{G} = \langle\hat{S}, U,
\hat{P}, \hat{r}, Z, \hat{O}, n, \gamma\rangle$. We can now write a Bellman equation for 
$\hat{G}$:
\begin{multline}
Q(\tau, u) = \sum_{\hat{s}} p(\hat{s} | \tau) \bigg[ \hat{r}(\hat{s},u)  ~+\\
\gamma \sum_{\tau', \hat{s}', u'} \hat{P}(\hat{s}' | \hat{s},u ) \pi(u' ,\tau') 
p(\tau'|\tau, \hat{s}',u) Q(\tau',u')  \bigg].
\label{eq:part_obs_bellman}
\end{multline}
Substituting back in the definitions of the quantities in $\hat{G}$, we arrive
at a Bellman equation of a form similar to \eqref{eq:fully_obs_bellman},
where the righthand side is multiplied by $\vpi_{-a}(\vu_{-a}|\vtau_{-a})$:
\begin{multline}
Q(\tau, u) = \sum_{\hat{s}} p(\hat{s} | \tau) \sum_{\vu_{-a}}
\vpi_{-a}(\vu_{-a}|\vtau_{-a}) \bigg[  r(s,\vu)  ~+\\ 
\gamma \sum_{\tau', \hat{s}', u'} P(s'|s,\vu) 
p(\vtau_{-a}'|\vtau_{-a},\vu_{-a},s') ~\cdot \\
\pi(u' ,\tau') p(\tau'|\tau, \hat{s}',u) Q(\tau',u')  \bigg].
\label{eq:part_obs_bellman2}
\end{multline}
This construction simply allows us to demonstrate the dependence of the Bellman
equation on the same nonstationary term $\vpi_{-a}(\vu_{-a}|s)$ in the
partially-observable case. However, unlike in the fully observable case, the
righthand side contains several other terms that indirectly depend on the
policies of the other agents and are to the best of our knowledge intractable.
Consequently, the importance ratio defined above,
$\frac{\vpi_{-a}^{t_r}(\vu_{-a}|s)}{\vpi_{-a}^{t_i}(\vu_{-a}|s)}$, is only an
approximation in the partially observable setting.

\subsection{Multi-Agent Fingerprints}

While importance sampling provides an unbiased estimate of the true objective,
it often yields importance ratios with large and unbounded
variance \cite{robert2004monte}. Truncating or adjusting the importance weights can reduce the variance but introduces bias.
Consequently, we propose an alternative method that embraces the
nonstationarity of multi-agent problems, rather than correcting for it.

The weakness of  IQL is that, by treating other agents as part of the
environment, it ignores the fact that such agents' policies are changing over
time, rendering its own Q-function nonstationary. This implies that the
Q-function could be made stationary if it conditioned on the policies of the
other agents.  This is exactly the philosophy behind \emph{hyper Q-learning}
\citep{tesauro2003extending}: each agent's state space is augmented with an
estimate of the other agents' policies computed via Bayesian inference.
Intuitively, this reduces each agent's learning problem to a standard,
single-agent problem in a stationary, but much larger, environment.

The practical difficulty of hyper Q-learning is that it increases the
dimensionality of the Q-function, making it potentially infeasible
to learn.  This problem is exacerbated in deep learning, when the other agents'
policies consist of high dimensional deep neural networks.  Consider a naive
approach to combining hyper Q-learning with deep RL that includes the weights of
the other agents' networks, $\vtheta_{-a}$, in the observation function. The new
observation function is then $O'(s) = \{O(s), \vtheta_{-a}\}$. The agent could
in principle then learn a mapping from the weights $\vtheta_{-a}$, and its own
trajectory $\tau$, into expected returns. Clearly, if the other agents are using
deep models, then $\vtheta_{-a}$ is far too large to include as input to the
Q-function.

However, a key observation is that, to stabilise experience replay, each agent
does not need to be able to condition on any possible  $\vtheta_{-a}$, but only
those values of  $\vtheta_{-a}$ that actually occur in its replay memory.  The
sequence of policies that  generated the data in this buffer can be thought of
as following a single, one-dimensional trajectory through the high-dimensional
policy space.  To stabilise experience replay, it should be sufficient if each
agent's observations disambiguate where along this trajectory the current
training sample originated from.

The question then, is how to design a low-dimensional \emph{fingerprint} that
contains this information. Clearly, such a fingerprint must be correlated with
the true value of state-action pairs given the other agents' policies. It should
typically vary smoothly over training, to allow the model to generalise
across experiences in which the other agents execute policies of varying quality
as they learn. An obvious candidate for inclusion in the fingerprint is the
training iteration number $e$. One potential challenge is that after policies
have converged, this requires the model to fit multiple fingerprints to the same
value, making the function somewhat harder to learn and more difficult to
generalise from.

Another key factor in the performance of the other agents is the
rate of exploration $\epsilon$. Typically an annealing schedule is set for $\epsilon$ such that it varies smoothly
throughout training and is quite closely correlated to performance. Therefore, we
further augment the input to the Q-function with $\epsilon$, such that the
observation function becomes $O'(s) = \{O(s), \epsilon, e \}$. Our results in Section \ref{sec:results} show that even
this simple fingerprint is remarkably effective.
%
%

\section{Experiments}
\label{sec:experiments}


In this section, we describe our experiments applying experience replay with 
fingerprints (XP+FP), with importance sampling (XP+IS), and with the combination (XP+IS+FP), to the StarCraft 
domain. We run experiments with both feedforward (FF) and recurrent (RNN) models, to test 
the hypothesis that in StarCraft recurrent models can use trajectory information 
to more easily disambiguate experiences from different stages of training.

\subsection{Decentralised StarCraft Micromanagement}

StarCraft is an example of a complex, stochastic environment whose dynamics cannot easily be simulated. This differs from standard multi-agent settings such as Packet World \citep{weyns2005packet} and simulated RoboCup \citep{hausknecht2016half}, where often entire episodes can be fully replayed and analysed. This difficulty is typical of real-world problems, and is well suited to the model-free approaches common in deep RL.
In StarCraft, \emph{micromanagement} refers to the subtask of controlling single or grouped units to move them around the map and fight enemy units.
In our multi-agent variant of StarCraft micromanagement, the centralised player 
is replaced by a set of agents, each assigned to one unit on the map. Each 
agent observes a subset of the map centred on the unit it controls, as 
shown in \reffig{problem-setting}, and must select from a restricted set of 
durative actions: 
\texttt{move[direction]}, \texttt{attack[enemy\_id]}, \texttt{stop}, and \texttt{noop}. 
During an episode, each unit is identified by a positive integer initialised on 
the first time-step.

\begin{figure*}[t!]
\centering
\subfigure[3v3 with recurrent networks]{
       \includegraphics[width=.45\textwidth]{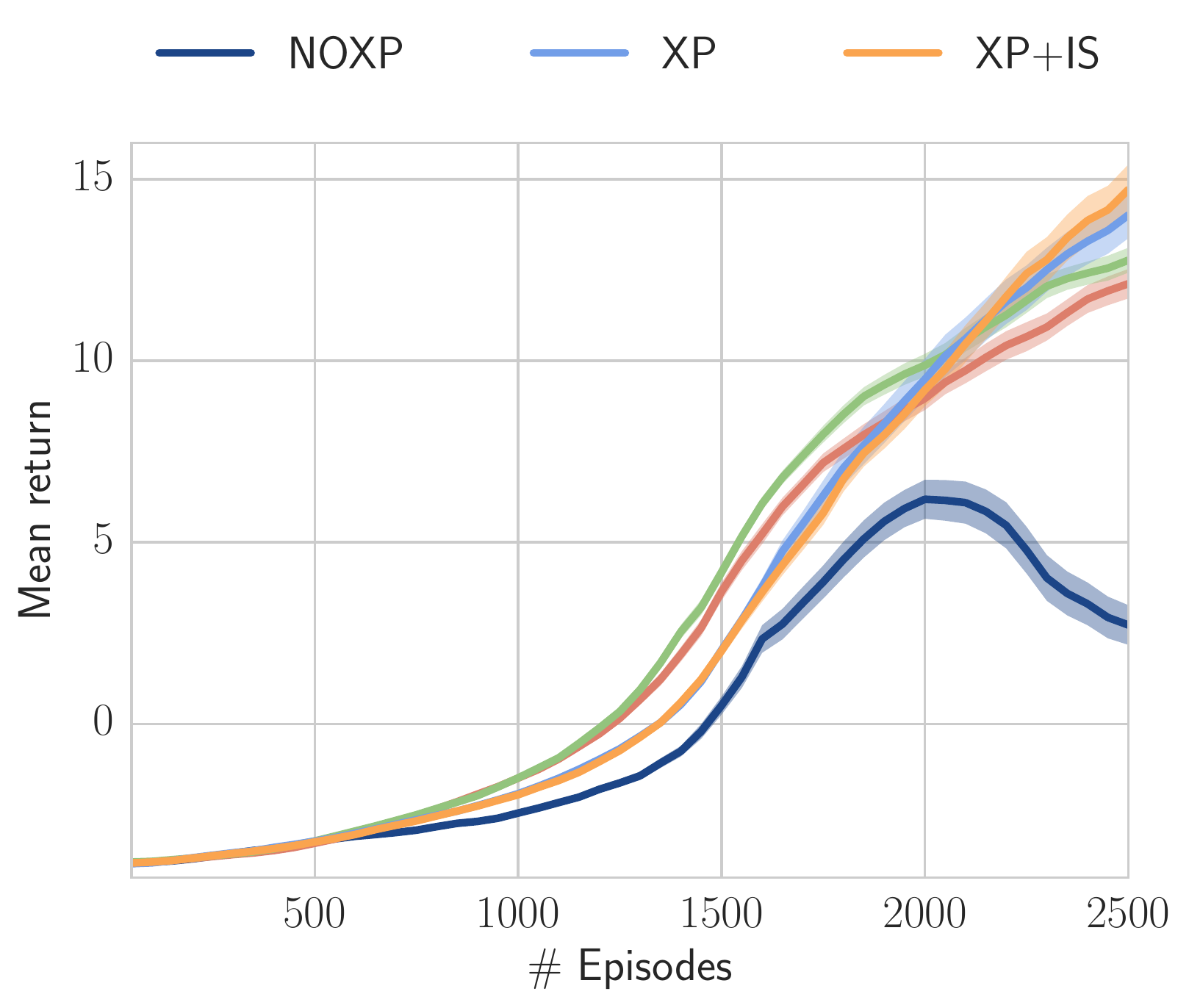}
}
\subfigure[3v3 with feed-forward networks]{
       \includegraphics[width=.45\textwidth]{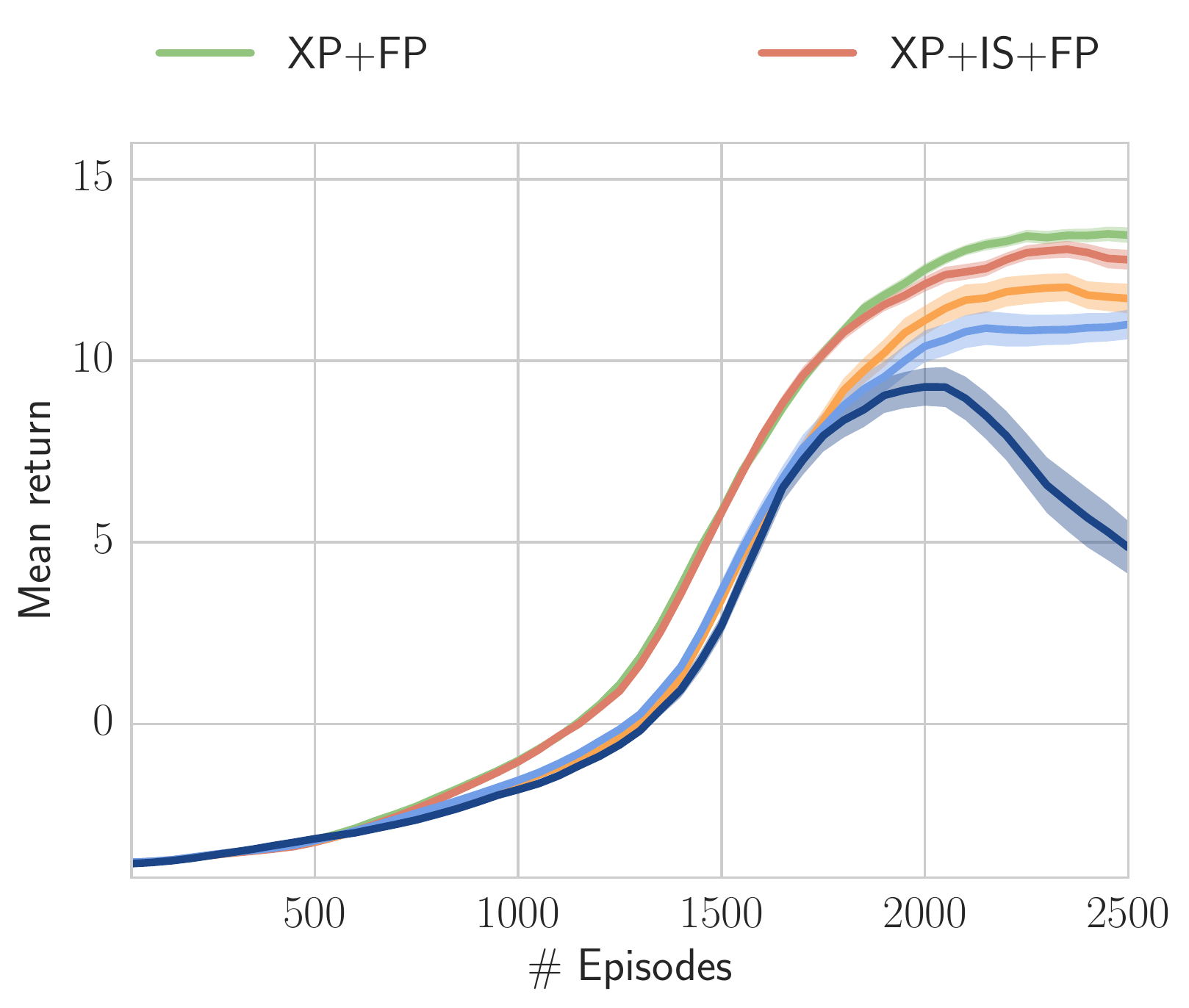}
}
\\
\subfigure[5v5 with recurrent networks]{
       \includegraphics[width=.45\textwidth]{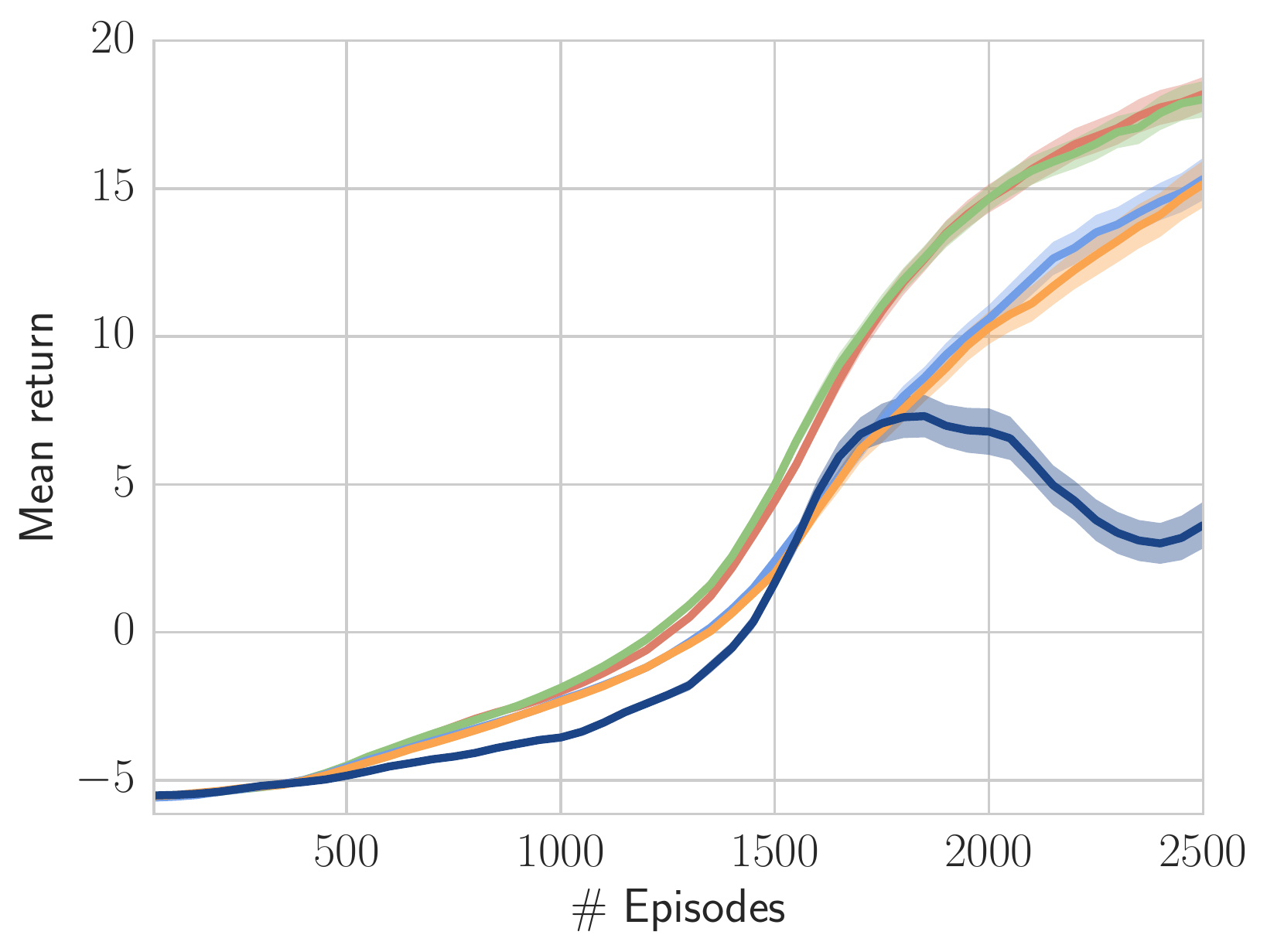}
     }
\subfigure[5v5 with feed-forward networks]{
       \includegraphics[width=.45\textwidth]{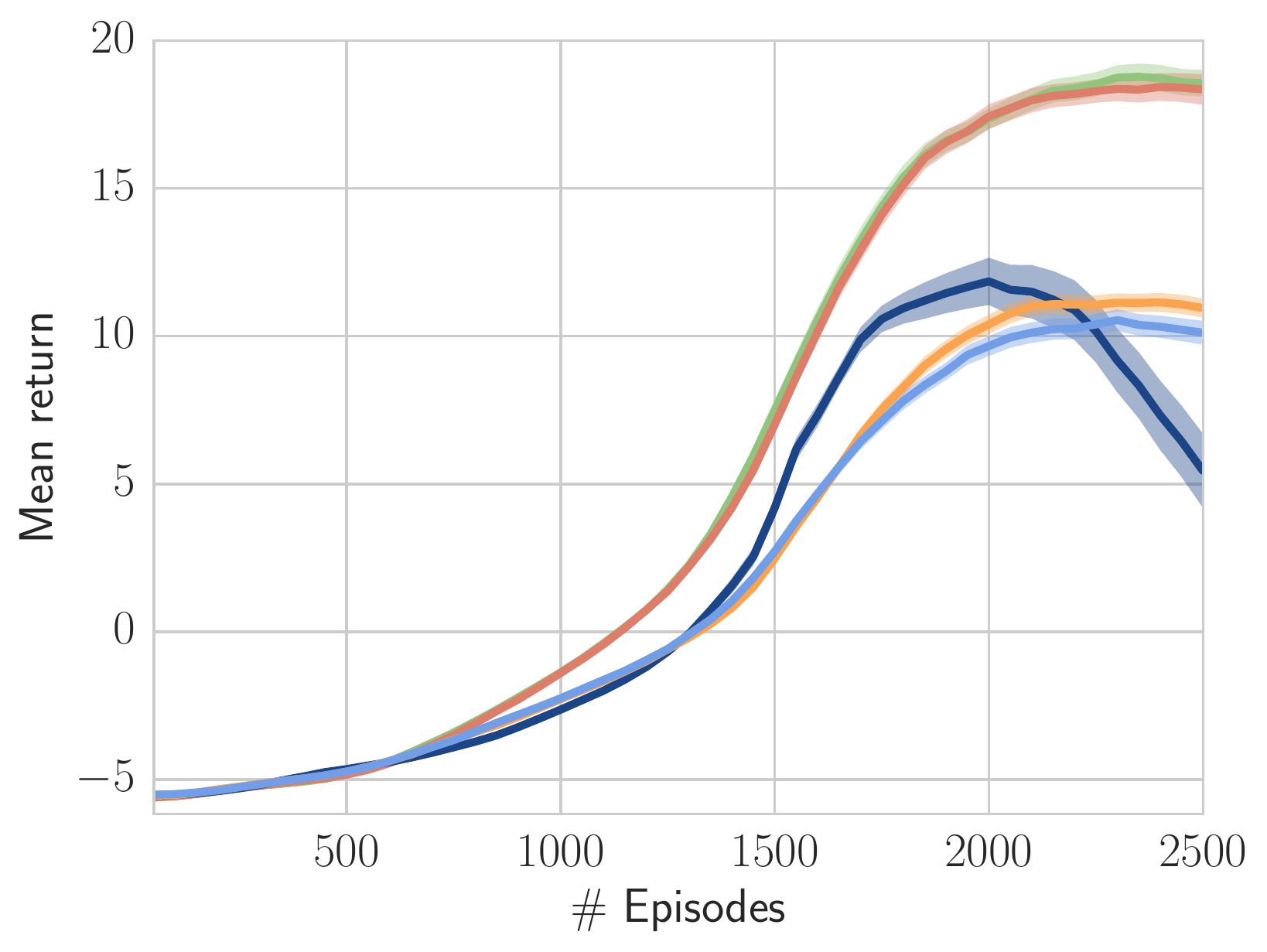}
}
\caption{Performance of our methods compared to the two baselines XP 
and NOXP, for both RNN and FF; (a) and (b) show the 3v3 setting, in 
which IS and FP are only required with feed-forward networks; (c) and 
(d) show the 5v5 setting, in which FP clearly improves performance 
over the baselines, while IS shows a small improvement only in the 
feedforward setting. Overall, the FP is a more effective method for 
resolving the nonstationarity and there is no additional benefit from 
combining IS with FP. Confidence intervals show one standard deviation 
of the sample mean.}
\label{fig:fig_1}
\end{figure*}

All units are \emph{Terran Marines}, ground units with a fixed range of fire about the length of four stacked units.  
Reward is the sum of the damage inflicted against opponent units during that timestep, with an additional terminal reward equal to the sum of the  health of all units on the team. This is a variation of a naturally arising battle signal, comparable with the one used by \citet{usunier2016episodic}.
A few timesteps after the agents are spawned, they are attacked by opponent units of the same type. Opponents are controlled by the game AI, which is set to attack all the time. We consider two variations: 3 marines vs 3 marines (m3v3), and 5 marines vs 5 marines (m5v5). Both of these require the agents to coordinate their movements to get the opponents into their range of fire with good positioning, and to focus their firing on each enemy unit so as to destroy them more quickly. Skilled human StarCraft players can typically solve these tasks.

We build our models in Torch7 \cite{torch}, and run our StarCraft experiments with TorchCraft \citep{synnaeve2016torchcraft}, a library that provides the functionality to enact the standard reinforcement learning step in \emph{StarCraft: BroodWar}, which we extend to enable multi-agent control.

\subsection{Architecture}

We use the recurrent DQN architecture described by \mbox{\citet{foerster2016learning}} with a few modifications. We do not consider communicating agents, so there are no message connections. As mentioned above, we use two different different models: one with a feed-forward model with two fully connected hidden layers, and another with a single-layer GRU. For both models, every hidden layer has 128 neurons.

We linearly anneal $\epsilon$ from 1.0 to 0.02 over 1500 episodes, and train the network for $e_{max} = 2500$ training episodes. In the standard training loop, we collect a single episode and add it to the replay memory at each training step. We sample batches of $\frac{30}{n}$ episodes uniformly from the replay memory and train on fully unrolled episodes. In order to reduce the variance of the multi-agent importance weights, we clip them to the interval $[0.01,2]$. We also normalise the importance weights by the number of agents, by raising them to the power of $\frac{1}{n -1}$. Lastly, we divide the importance weights by their running average in order to keep the overall learning rate constant. All other hyperparameters are identical to \citet{foerster2016learning}.

\section{Results}
\label{sec:results}

In this section, we present the results of our StarCraft experiments, summarised
in Figure \ref{fig:fig_1}. Across all tasks and models, the baseline without experience replay (NOXP) performs
poorly. Without the diversity in trajectories provided by experience replay, NOXP overfits to the greedy policy once
$\epsilon$ becomes small. When exploratory actions do occur,
agents  visit areas of the state space that have not had their $Q$-values
updated for many iterations, and bootstrap off of values which have become
stale or distorted by updates to the $Q$-function elsewhere. This effect can harm or destabilise the policy. 
With a recurrent model,
performance simply degrades, while in the feed-forward case, it
begins to drop significantly later in training. We hypothesise that full
trajectories are inherently more diverse than single observations, as they
include compounding chances for exploratory actions. Consequently, it is easier
to overfit to single observations, and experience replay is more essential for a
feed-forward model.

With a naive application of experience replay (XP), the model tries to simultaneously learn a best-response policy to every
historical policy of the other agents. Despite the 
nonstationarity, the stability of experience replay enables XP to
outperform NOXP in each case. However, due to limited
disambiguating information, the model cannot appropriately
account for the impact of any particular policy of the other agents, or keep
track of their current policy. The experience replay is therefore used
inefficiently, and the model cannot generalise properly from experiences early
in training.

\subsection{Importance Sampling}
The importance sampling approach (XP+IS) slightly outperforms XP when using
feed-forward models. While mathematically sound in the fully observable case, XP+IS is only
approximate for our partially observable problem, and runs into practical
obstacles. Early in training, the importance weights are relatively well behaved
and have low variance. However, as $\epsilon$ drops, the importance ratios become
multi-modal with increasing variance. The large majority of importance
weights are less than or equal to $\epsilon(1-\epsilon) \approx \epsilon$, so
few experiences contribute strongly to learning. In a setting that does not
require as strongly deterministic a policy as StarCraft, $\epsilon$ could be
kept higher and the variance of the importance weights would be lower.


\subsection{Fingerprints}
Our results show that  the simple fingerprint of adding $e$ and
$\epsilon$ to the observation (XP+FP) dramatically improves performance for the
feed-forward model. This fingerprint provides sufficient disambiguation for the model to track the quality of the other agents'
policies over the course of training, and make proper use of the experience
buffer. The network still sees a diverse array of input states across which to generalise but is able to modify its predicted value in accordance with the known
stage of training. 

Figure \ref{fig:fig_1} also shows that there is no extra benefit from combining importance sampling with fingerprints (XP+IS+FP). This makes sense given that the two approaches both address the same problem of nonstationarity, albeit in different ways.

\begin{figure}[ht]
	\centering
	\includegraphics[width=\columnwidth]{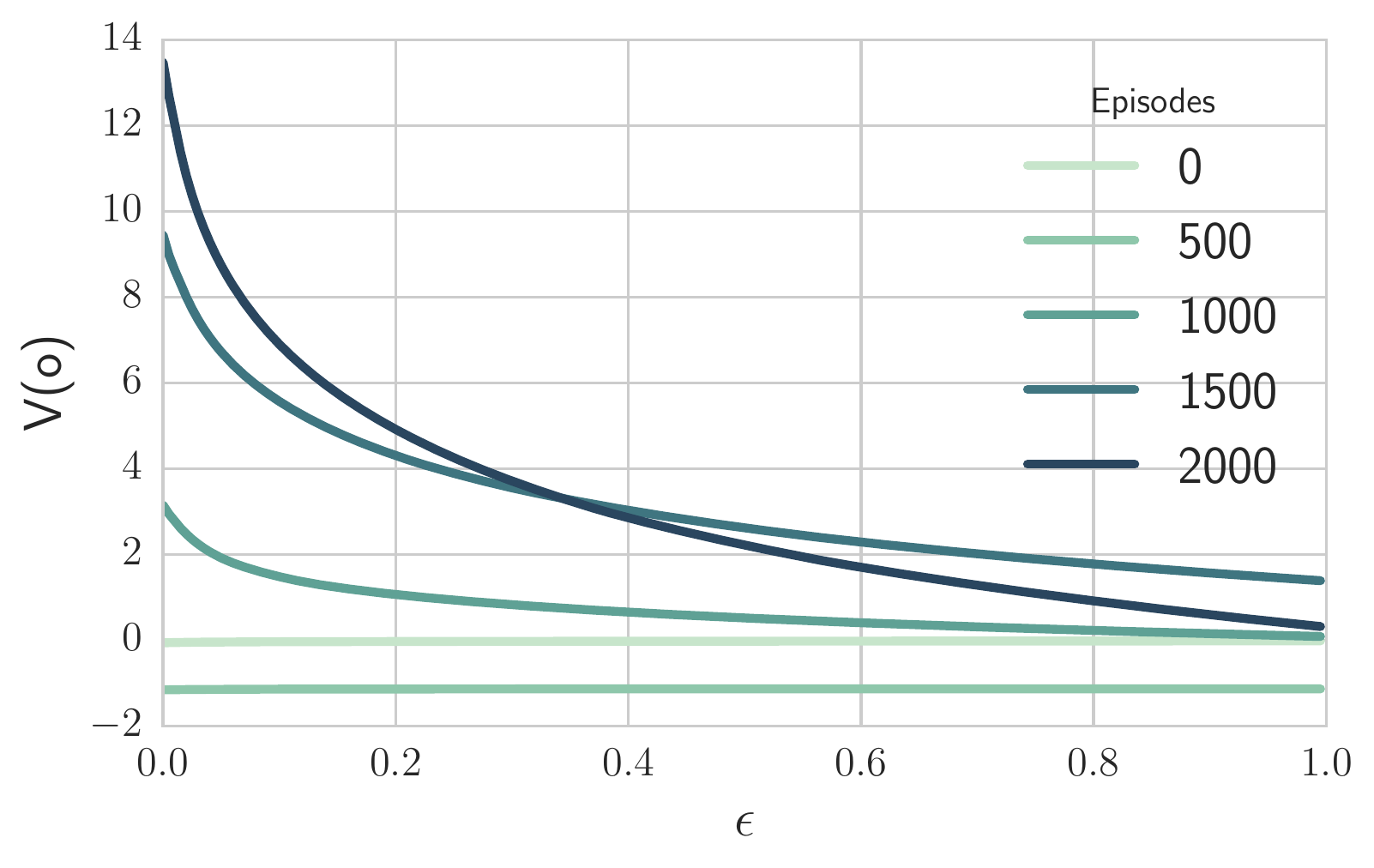}
	\caption{Estimated value of a single initial observation with different 
    $\epsilon$ in its fingerprint input, at different stages of training. The network learns to smoothly vary its value estimates across different stages of training.}
	\label{fig:fp_sweep}
\end{figure}

Figure \ref{fig:fp_sweep}, which shows the estimated value for XP+FS of a
single initial state observation with different $\epsilon$ inputs, demonstrates that the
network learns to smoothly vary its value estimates across different stages of training,
correctly associating high values with the low $\epsilon$ seen later in
training. This approach allows the model to generalise between best responses to
different policies of other agents. In effect, a larger dataset is available in
this case than when using importance sampling, where most experiences are
strongly discounted during training. The fingerprint enables the transfer
of learning between diverse historical experiences, which can significantly
improve performance.

\subsection{Informative Trajectories}
\begin{figure}[ht]
\centering
\subfigure[]{
	\includegraphics[width=.45\columnwidth]{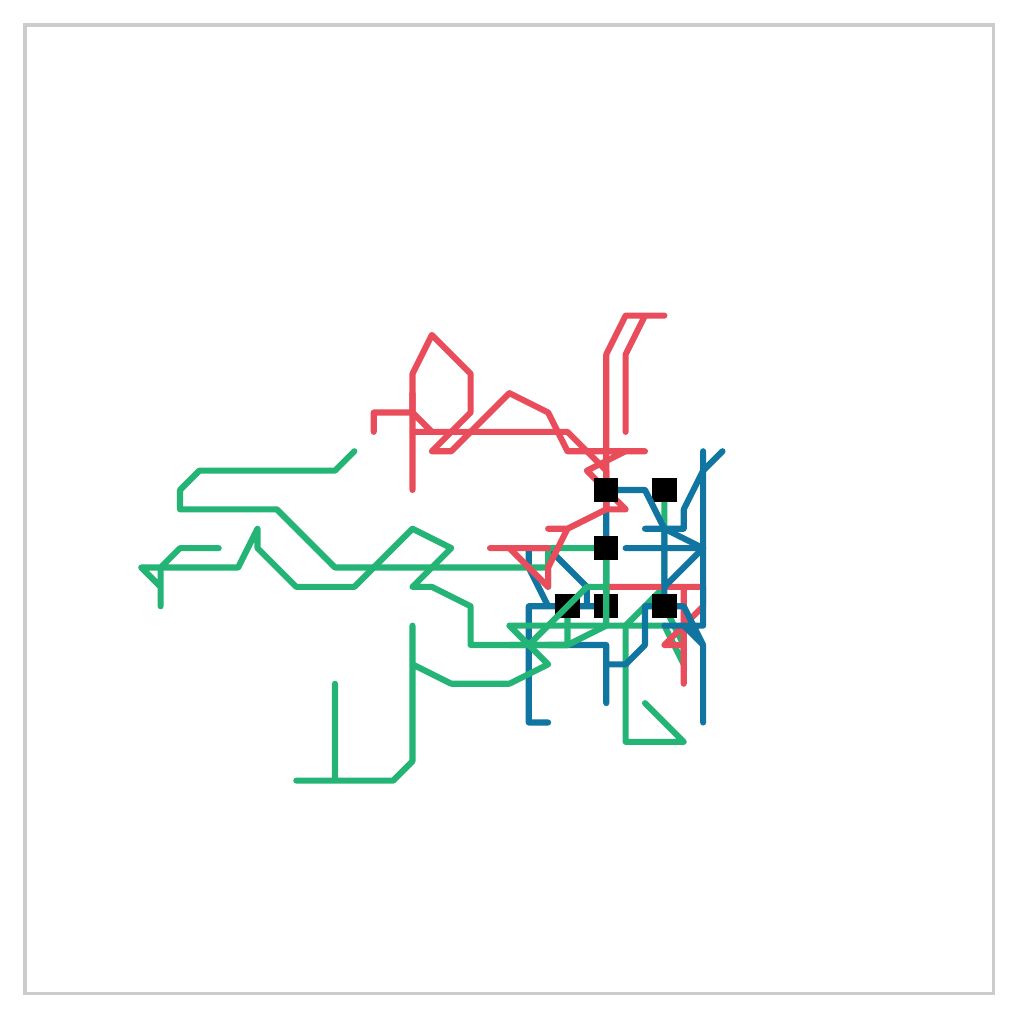}
}
\subfigure[]{
	\includegraphics[width=.45\columnwidth]{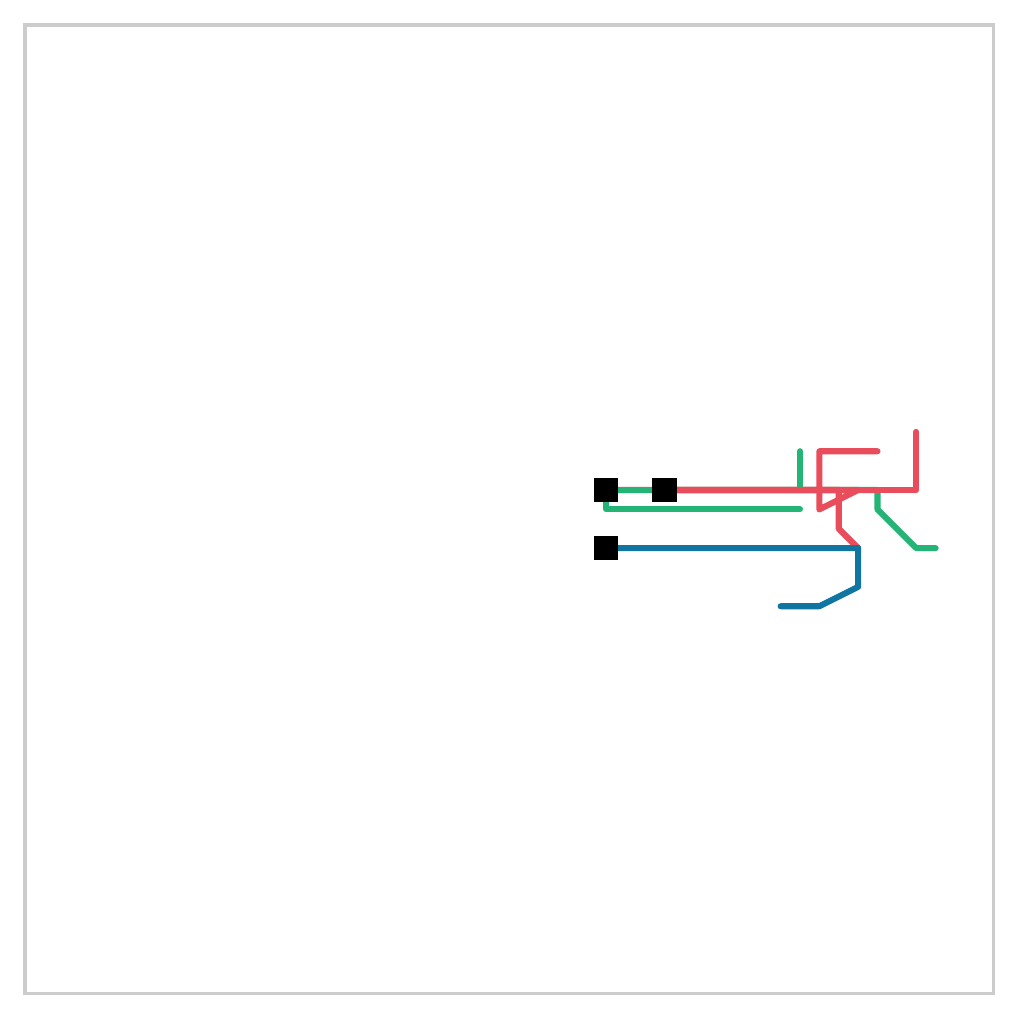}
}
\subfigure[]{
\includegraphics[width=.45\columnwidth]{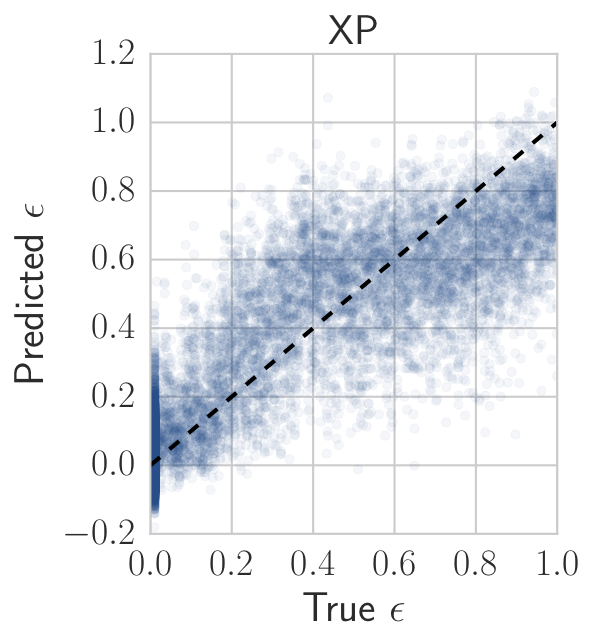}
}
\subfigure[]{
	\includegraphics[width=.45\columnwidth]{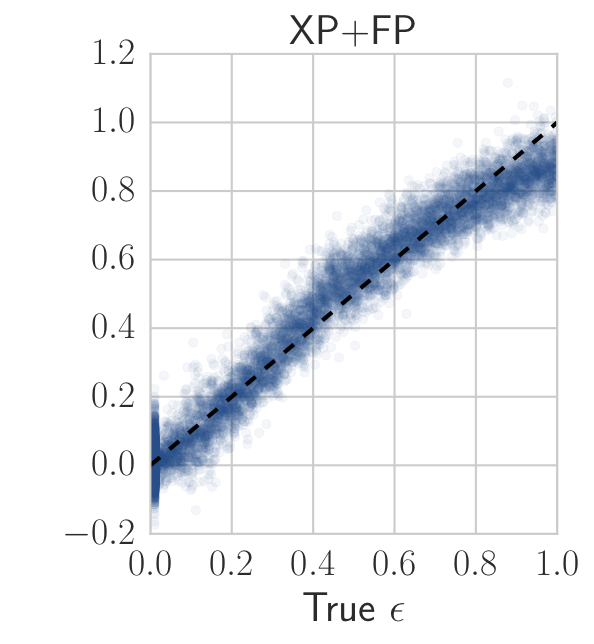}
}

\caption{(upper) Sampled trajectories of agents, from the beginning (a) and end
(b) of training. Each agent is one colour and the starting points are marked as
black squares. (lower) Linear regression predictions of $\epsilon$ 
from the
hidden state halfway through each episode in the replay buffer: (c) with only
XP, the hidden state still contains disambiguating information drawn 
from the
trajectories, (d) with XP+FP, the hidden state is more informative about the
stage of training.} 
\label{fig:traj}
\end{figure}

When using recurrent networks, the performance gains of XP+IS and XP+FP are not
as large; in the 3v3 task, neither method helps. The reason is that, in
StarCraft, the observed trajectories are significantly informative about the
state of training, as shown in Figurea \ref{fig:traj}a and \ref{fig:traj}b. For
example, the agent can observe that it or its allies have taken many seemingly
random actions, and infer that the sampled experience comes from early in
training. This is a demonstration of the power of recurrent architectures in
sequential tasks with partial observability: even without explicit additional
information, the network is able to partially disambiguate experiences from
different stages of training. To illustrate this, we train a linear model to
predict the training $\epsilon$ from the hidden state of the recurrent model.
Figure \ref{fig:traj}c shows a reasonably strong predictive accuracy even for a
model trained with XP but no fingerprint, indicating that disambiguating
information is indeed kept in the hidden states. However, the hidden states of a
recurrent model trained with a fingerprint (Figure \ref{fig:traj}d) are even more informative.

\section{Conclusion}
\label{sec:conclusion}

This paper
proposed two methods for stabilising experience replay in deep multi-agent reinforcement learning: 1) using a multi-agent variant
of importance sampling to naturally decay obsolete data and 2)  conditioning
each agent's value function on a fingerprint that disambiguates the age
of the data sampled from the replay memory.  Results on a challenging
decentralised variant of StarCraft unit micromanagement  confirmed that
these methods enable the successful combination of experience replay with multiple agents.
In the future, we would like to apply these methods to a broader range of nonstationary training problems, such as classification on changing data, and extend them to multi-agent actor-critic methods.

\newpage
\section*{Acknowledgements} 
This project has received funding from the European Research Council (ERC) under
the European Union’s Horizon 2020 research and innovation programme (grant
agreement \#637713). This work was also supported by the Oxford-Google DeepMind
Graduate Scholarship, the Microsoft Research PhD Scholarship Program, EPSRC AIMS
CDT grant EP/L015987/1, ERC grant ERC-2012-AdG 321162-HELIOS, EPSRC grant
Seebibyte EP/M013774/1 and EPSRC/MURI grant EP/N019474/1.
Cloud computing GPU resources were provided through a Microsoft Azure for
Research award.
We thank Nando de Freitas, Yannis Assael, and Brendan Shillingford for the
helpful comments and discussion. We also thank Gabriel Synnaeve, Zeming Lin, and
the rest of the TorchCraft team at FAIR for all the help with the interface.

\bibliography{starcomm,mrl}
\bibliographystyle{include/icml2017}

\end{document}